# A Novel Reservoir Computing Framework for Chaotic Time Series Prediction Using Time Delay Embedding and Random Fourier Features

S. K. Laha

Advanced Design and Analysis Group
CSIR-Central Mechanical Engineering Research Institute
MG Avenue, Durgapur, West Bengal, PIN-713209, India

**Abstract:**

Forecasting chaotic time series requires models that can capture the intrinsic geometry of the underlying attractor while remaining computationally efficient. We introduce a novel reservoir computing (RC) framework that integrates time-delay embedding with Random Fourier Feature (RFF) mappings to construct a dynamical reservoir without the need for traditional recurrent architectures. Unlike standard RC, which relies on high-dimensional recurrent connectivity, the proposed RFF-RC explicitly approximates nonlinear kernel transformations that uncover latent dynamical relations in the reconstructed phase space. This hybrid formulation offers two key advantages: (i) it provides a principled way to approximate complex nonlinear interactions among delayed coordinates, thereby enriching the effective dynamical representation of the reservoir, and (ii) it reduces reliance on manual reservoir hyperparameters such as spectral radius and leaking rate. We evaluate the framework on canonical chaotic systems-the Mackey-Glass equation, the Lorenz system, and the Kuramoto-Sivashinsky equation. This novel formulation demonstrates that RFF-RC not only achieves superior prediction accuracy but also yields robust attractor reconstructions and long-horizon forecasts. These results show that the combination of delay embedding and RFF-based reservoirs reveals new dynamical structure by embedding the system in an enriched feature space, providing a computationally efficient and interpretable approach to modeling chaotic dynamics.

# 1. INTRODUCTION

Reservoir computing (RC) is a Recurrent Neural Network (RNN) computing paradigm designed to handle time-series data efficiently. Unlike traditional RNNs, where training is computationally intensive, RC utilizes a randomly initialized and fixed recurrent layer (the reservoir) while only training the output weights. This significantly reduces computational complexity while maintaining predictive power. The key theoretical basis for RC is the echo state property, which ensures that past inputs influence the reservoir's state in a stable manner, making RC particularly effective for modeling dynamical systems.

Jaeger [1] introduced the Echo State Network (ESN), demonstrating how a randomly initialized recurrent network with fixed weights could effectively process sequential data. This work established the foundational principles of RC. Lukoševičius and Jaeger [2, 3] provided a comprehensive review of RC methodologies, detailing the techniques for designing reservoirs and training output layers. Gauthier et al. [4] examined RC's efficiency in learning deterministic dynamical systems, emphasizing its strengths in working with small datasets and linear optimization. Bollt [5] analyzed RC's success from a dynamical systems perspective, linking it to vector autoregressive models and dynamic mode decomposition (DMD). He provided theoretical insights into its success.

Data-driven dynamical systems methodologies employ computational inference and statistical/machine learning frameworks to model nonlinear dynamics, attractor reconstruction, and forecast chaotic and spatiotemporal phenomena [6]. Core paradigms encompass system identification via sparse regression, dimensionality reduction of high-dimensional phase spaces, and neural operators approximating solution functions of partial differential equations. Brunton et al. [7] introduced Sparse Identification of Nonlinear Dynamical Systems (SINDy), a method that extracts governing equations using sparse regression, that enables interpretable modeling of complex systems. Schmid [8] reviewed Dynamic Mode Decomposition (DMD), a technique that extracts dominant spatiotemporal structures (DMD modes) and their evolution over time, making it widely used in fluid dynamics, neuroscience, etc. Kutz et al. [9] provided a detailed mathematical foundation for DMD and its applications in data-driven modeling. Vlachas et al. [10] used Long Short-Term Memory (LSTM) networks to predict chaotic systems, demonstrating their effectiveness in reduced-order modeling. Gilpin [11] proposed an autoencoder-based method to reconstruct strange attractors from time-series data. Raissi et al. [12] developed Physics-Informed Neural

Networks (PINNs), which incorporate physical laws into machine learning frameworks to solve PDEs accurately. Li et al. [13] introduced Fourier Neural Operators, a method for solving parametric partial differential equations efficiently by learning mappings between function spaces.

RC has been widely applied in predicting chaotic and spatio-temporal systems [14-22]. Pathak et al. [14] demonstrated RC's ability in predicting large spatio-temporally chaotic systems (e.g., Kuramoto-Sivashinsky equation). Pathak et al. [15] also proposed a hybrid forecasting method combining RC with knowledge-based models. This approach enhances prediction accuracy and extends forecast horizons by leveraging both data-driven and mechanistic strengths. Chattopadhyay et al. [17] compared RC, ANN, and LSTM for multi-scale Lorenz 96 system forecasting, showing RC's superior short-term accuracy. Pandey and Schumacher [18] applied RC to model two-dimensional turbulent convection. Using proper orthogonal decomposition (POD) for dimensionality reduction, RC accurately predicts dominant mode evolution, aligning with direct numerical simulations. Kobayashi et al. [20] analyzed RC from a dynamical systems perspective, assessing its reconstruction of unstable fixed points, periodic orbits, and other features. It accurately predicts laminar lasting time distributions in fluid flows. Chen et al. [21] introduced a calibrated RC method with feedback to enhance reconstruction accuracy and length, significantly improving performance on chaotic time series. Zimmermann and Parlitz [22] applied RC to model spatio-temporal dynamics in excitable media, such as cardiac tissue, achieving robust predictions even in noisy conditions.

Recent studies have focused on enhancing RC's structure for better efficiency and accuracy [23-33]. Gallicchio et al. [23] explored deep RC with stacked reservoirs which shows improvements in memory capacity and temporal representation. Gauthier et al. [24] proposed next-generation RC that eliminates random matrices and reduces hyper-parameter tuning, achieving faster training and better generalization based on nonlinear vector auto-regression (NVAR) and delay embedding. Liu et al. [25] incorporated physics constraints into RC to improve predictions in chaotic fluid flow models. Other state-of-the-art echo state network models include intrinsic plasticity [27-29], small world topology [30], metaheuristic algorithm optimized ESN [31-32], Deep Fuzzy ESN [33] etc.

Kernel methods are a class of machine learning techniques that project input data into a high-dimensional space, enabling efficient separation of nonlinear patterns. However, exact kernel computations can be expensive. Random Fourier Features (RFF) provides a computationally efficient approximation of shift-invariant kernel functions (such as Gaussian/RBF kernel). Rahimi and Recht [34] introduced RFF as a technique to approximate shift-invariant kernels, which significantly speeds up kernel methods. Li et al. [35] analyzed the theoretical properties of RFF, improving feature count bounds for kernel ridge regression. Avron et al. [36] analyzed RFF from a spectral matrix approximation perspective, providing tight bounds on feature counts for kernel ridge regression. It discusses statistical guarantees and computational trade-offs. Sutherland and Schneider [37] examined the variance and performance trade-offs of different RFF variants, refining approximation bounds for kernel methods.

Several comprehensive surveys [38-41] highlight RC's evolution and emerging applications. Tanaka et al. [38] reviewed physical implementations of RC, showcasing energy-efficient hardware designs. Yan et al. [39] provided an overview of RC's research landscape, discussing ongoing challenges and potential advancements. Bai et al. [40] explored RC's role in mobile edge intelligence, covering applications in IoT and communication networks while Zhang et al. [41] reviewed early RC models, then state-of-the-art RC models and offers perspective on interaction of RC, cognitive neuroscience and evolution.

Just as traditional reservoir computing (RC) transforms an input $u(n)$ into a high-dimensional state representation $x(n)$ to capture complex dynamics, RFF approximates kernel functions by mapping inputs into a randomized high-dimensional feature space. This ensures that nonlinear relationships in the data are preserved while enabling efficient computations. We developed a novel reservoir computing (RC) method for predicting chaotic time series by combining delay embedding and Random Fourier Features (RFF) to efficiently capture temporal dependencies and nonlinear dynamics. Using delay embedding, we encode historical data into composite vectors to reconstruct the system's state space, which are then mapped into a high-dimensional, nonlinear feature space via RFF's random projections approximating a Gaussian kernel. A multi-output Ridge regression model, with $L_2$ regularization, is trained on these features to forecast the next time step. Multi-step ahead predictions are achieved by iteratively feeding predictions back into updated delay vectors. This approach simplifies hyper-parameter tuning compared to traditional RC and delivers high accuracy on benchmark chaotic systems like Mackey-Glass, Lorenz63, and Kuramoto-Sivashinsky by offering a computationally efficient solution for chaotic time series forecasting.

## 2. METHOD

### 2.1. ECHO STATE NETWORK

An Echo State Network (ESN) is a type of reservoir computing model used for time series prediction and dynamical system modeling. It is a special type of recurrent neural network (RNN) where only the output weights are trained, while the recurrent connections in the hidden layer (reservoir) remain fixed after initialization. The ESN is designed to have the echo state property, ensuring that past inputs fade over time and do not dominate future states.

The ESN consists of three main layers: input, reservoir, and output. Given a time-dependent $d$-dimensional input $\mathbf{u}(t) \in \mathbb{R}^d$, the reservoir state $\mathbf{x}(t) \in \mathbb{R}^N$ updates according to

$$\mathbf{x}(t+1) = (1-\alpha)\mathbf{x}(t) + \alpha f(W_r \mathbf{x}(t) + W_{\text{in}} \mathbf{u}(t) + b) \tag{1}$$

where:
- $W_{\text{in}} \in \mathbb{R}^{N \times d}$ is the randomly initialized input weight matrix
- $W_r \in \mathbb{R}^{N \times N}$ is the reservoir weight matrix, typically sparse and scaled
- $b \in \mathbb{R}^N$ is a bias term
- $f$ is a nonlinear activation function, such as tanh,
- $\alpha \in (0,1]$ is the leaking rate.

The output is computed as

$$\mathbf{y}(t) = W_{\text{out}} \mathbf{x}(t) \tag{2}$$

where $W_{\text{out}} \in \mathbb{R}^{m \times N}$ is the trainable output weight matrix, learned using a simple linear regression (e.g., ridge regression or least squares). Unlike traditional RNNs, only $W_{\text{out}}$ is optimized, making ESNs computationally efficient.

To ensure the echo state property, the spectral radius $\rho(W_r)$ (largest absolute eigenvalue of $W_r$) is typically set to be less than 1.

### 2.2. PRESENT METHODOLOGY

Consider a $d$-dimensional time series $\mathbf{u}(t) = [u_1(t), u_2(t), \ldots, u_d(t)]^T$ observed at discrete time steps $t = 1,2,\ldots,N$, where the objective is to forecast the next state $\mathbf{u}(t+1)$ using the past data. For example, in the Lorenz system, $d = 3$ with $u_1(t) = x(t), u_2(t) = y(t)$ and $u_3(t) = z(t)$ represents the system's state variables.

Random Fourier Features (RFF) [34] is a powerful technique rooted in Bochner's theorem, which asserts that any continuous, shift-invariant kernel (such as the Gaussian kernel $k(\mathbf{x}, \mathbf{y}) = \exp(-\|\mathbf{x} - \mathbf{y}\|^2/(2\sigma^2))$) can be represented as the Fourier transform of a positive measure. By approximating this transform through random projections, RFF maps input vectors into a finite-dimensional space where the inner products closely approximate the values of the Gaussian kernel, thereby enabling the capture of non-linear relationships in data. In this forecasting framework, RFF transforms delay vectors into a higher-dimensional feature space, setting the stage for effective modeling via Ridge regression. To embed temporal dependencies [42-43], we first form delay vectors for each variable $u_i(t) \in \mathbb{R}$ by taking its most recent $k$ observations, yielding

$$\mathbf{u}_i^{\text{delay}}(t) = [u_i(t), u_i(t-1), \ldots, u_i(t-k+1)]^T \tag{3}$$

By concatenating these $d$ individual delay vectors, we obtain the composite delay vector

$$\mathbf{u}^{\text{delay}}(t) = [\mathbf{u}_1^{\text{delay}}(t)^T, \mathbf{u}_2^{\text{delay}}(t)^T, \ldots, \mathbf{u}_d^{\text{delay}}(t)^T]^T \in \mathbb{R}^{dk} \tag{4}$$

which encapsulates the system's state over a window of $k$ time steps. To capture complex non-linear dynamics, we then map these delay vectors into an $m$-dimensional feature space using RFF. This is achieved by generating a random weight matrix $\mathbf{W} \in \mathbb{R}^{dk \times m}$ with entries $w_{ij} \sim \mathcal{N}(0, \frac{1}{\sigma_{\text{rff}}^2})$ (where $\sigma_{\text{rff}}$ controls the kernel width) and a random bias vector $\mathbf{b} \in \mathbb{R}^m$ with entries $b_j$ drawn uniformly from $[0, 2\pi]$. The non-linear mapping is then given by

$$\phi\left(\mathbf{u}^{\text{delay}}(t)\right) = \sqrt{\frac{2}{m}} \cos(\mathbf{W}^T \mathbf{u}^{delay}(t) + \mathbf{b}) \tag{5}$$

This yields an $m$-dimensional representation where the inner products approximate those of the Gaussian kernel. Stacking these transformed vectors for $t = k, k+1, \ldots, N-1$ forms the feature matrix $\Phi \in \mathbb{R}^{(N-k) \times m}$ while the corresponding targets are the next states $\mathbf{y}(t) = \mathbf{u}(t+1)$ arranged as rows in the target matrix $Y \in \mathbb{R}^{(N-k) \times d}$.

With these matrices, we set up a multi-output Ridge regression problem to find a linear mapping defined by the weight matrix $\mathbf{W}_{\text{ridge}} \in \mathbb{R}^{m \times d}$ and bias vector $\mathbf{b}_{\text{ridge}} \in \mathbb{R}^d$ that minimizes the regularized loss function

$$\min_{\mathbf{W}_{\text{ridge}}, \mathbf{b}_{\text{ridge}}} \left\| Y - \left(\Phi \mathbf{W}_{\text{ridge}} + \mathbf{1}\mathbf{b}_{\text{ridge}}^T\right) \right\|_F^2 + \lambda_{\text{reg}} \|\mathbf{W}_{\text{ridge}}\|_F^2 \tag{6}$$

where $\lambda_{\text{reg}} > 0$ is the regularization parameter and $\mathbf{1}$ is a column vector of ones. To solve this optimization, the target matrix $Y$ is centered by subtracting its mean $\bar{Y}$, and the design matrix $\Phi$ is also centered so that that $Y_c = Y - \bar{Y}$ and $\Phi_c = \Phi - \bar{\Phi}$.

The optimal weights are then obtained via the closed-form solution

$$\mathbf{W}_{\text{ridge}} = (\Phi_c^T \Phi_c + \lambda I)^{-1} \Phi_c^T Y_c \tag{7}$$

where $I$ denotes the identity matrix. And the bias term can be obtained as

$$\mathbf{b}_{\text{ridge}} = Y^T - \bar{\Phi} \mathbf{W}_{\text{ridge}} \tag{8}$$

For multi-step forecasting, an iterative approach is used: starting with the last known delay vector, we predict the next state using

$$\hat{\mathbf{y}}(t) = \phi\left(\mathbf{u}^{\text{delay}}(t)\right) \mathbf{W}_{\text{ridge}} + \mathbf{b}_{\text{ridge}} \tag{9}$$

then update the delay vector by discarding its oldest observation and appending the new prediction. This process is repeated iteratively to generate a sequence of forecasts $\hat{\mathbf{u}}(t+1), \hat{\mathbf{u}}(t+2), \ldots$ that effectively capture the system's evolving dynamics.

## 3. RESULTS

In dynamical systems governed by differential equations $\dot{\mathbf{x}} = f(\mathbf{x})$, the flow of states $\mathbf{x}(t)$ evolves continuously, where the stroboscopic solutions given by

$$\mathbf{x}(t + \tau) = \mathbf{x}(t) + \int_t^{t+\tau} f(\mathbf{x}(s)) ds. \tag{10}$$

Learning such dynamics demands models that efficiently approximate complex state evolutions across varied scales. Here, in this context, we propose a Random Fourier Feature-based Reservoir Computing (RFF-RC) framework. We demonstrate its efficacy on three systems: the chaotic Lorenz63 (3D), the delay-driven Mackey-Glass equation (1D), and the high-dimensional Kuramoto-Sivashinsky equation, showing that RFF-RC reliably learns stroboscopic state transitions $\mathbf{x}(t) \to \mathbf{x}(t+\tau)$.

## 3.1 Prediction of Lorenz Equation Using Random Fourier Features-Based Reservoir Computing (RFF-RC)

The Lorenz63 system is a set of three coupled, nonlinear differential equations originally formulated to model atmospheric convection but later recognized as a fundamental example of deterministic chaos [44]. The equations are given by:

$$\frac{dx}{dt} = \sigma(y - x), \frac{dy}{dt} = x(\rho - z) - y, \frac{dz}{dt} = xy - \beta z \tag{11}$$

where $\sigma = 10$, $\rho = 28$, and $\beta = \frac{8}{3}$ are the standard parameters that produce chaotic behavior. The solutions of this system exhibit sensitive dependence on initial conditions, leading to the characteristic butterfly-shaped attractor. In this study, we generated a dataset of 4000 time steps using the fourth-order Runge-Kutta method with a step size of 0.025, allocating 60% for training, 20% for testing, and 20% for validation. The RFF-RC model was optimized through grid search over key hyper-parameters, including the delay embedding dimension ($k$), the number of Random Fourier Features ($m$), the regularization parameter ($\lambda_{\text{reg}}$), and the kernel width ($\sigma_{\text{rff}}$). The optimal values obtained were $k = 5$, $m = 3000$, $\lambda = 10^{-6}$, and $\sigma_{\text{rff}} = 2.0$. The effectiveness of the optimized model is evident from the results shown in Fig. (1). The first plot in Fig. (1) presents a one-step-ahead prediction of the Lorenz system, comparing ground truth (blue) with predicted values (magenta) for the $x, y, z$ variables. The figure demonstrates near-perfect overlap and low Normalized Root Mean Squared Error (NRMSE) values of $4.08 \times 10^{-5}$, $1.27 \times 10^{-4}$, and $1.19 \times 10^{-4}$ for $x, y$, and $z$, respectively on the test data-set. The second middle plot (in Fig. (1)) displays the 3D phase space trajectories of the Lorenz system, where the predicted trajectories closely follow the true Lorenz attractor structure during both the training and testing phases, confirming the model's ability to reconstruct chaotic dynamics. The third plot evaluates the model's multi-step-ahead prediction capabilities and error propagation over time. Initially, the predicted trajectory aligns well with the true system dynamics, but errors accumulate as the prediction horizon extends due to the intrinsic sensitivity of chaotic systems. The right panel of the third plot quantifies this error growth, showing that the model maintains reliable forecasting up to approximately five Lyapunov times before divergence becomes significant. These results highlight the efficacy of the RFF-RC approach in capturing and predicting chaotic behavior, thus, providing a computationally efficient alternative to traditional Echo State Networks (ESNs) while reducing reliance on extensive hyper-parameter tuning.

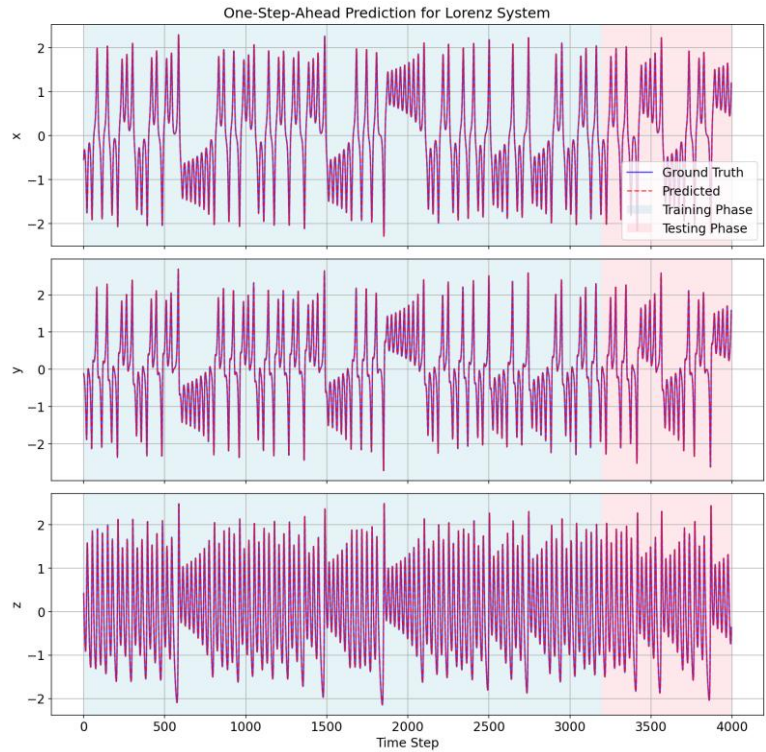

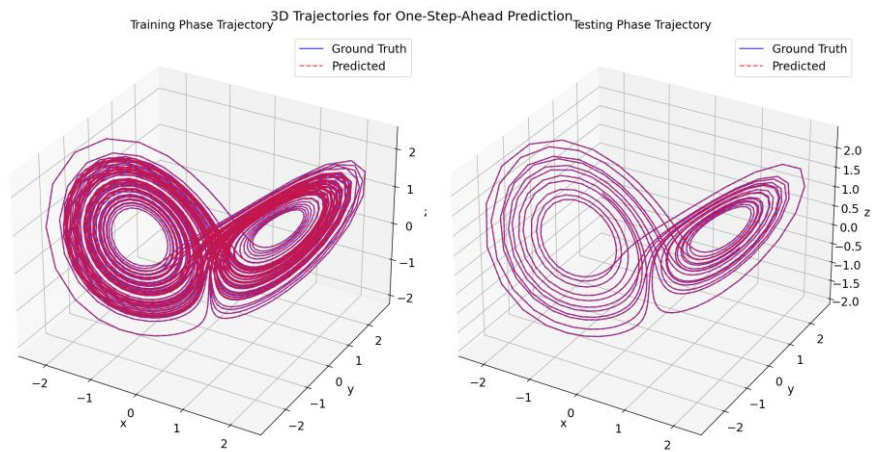

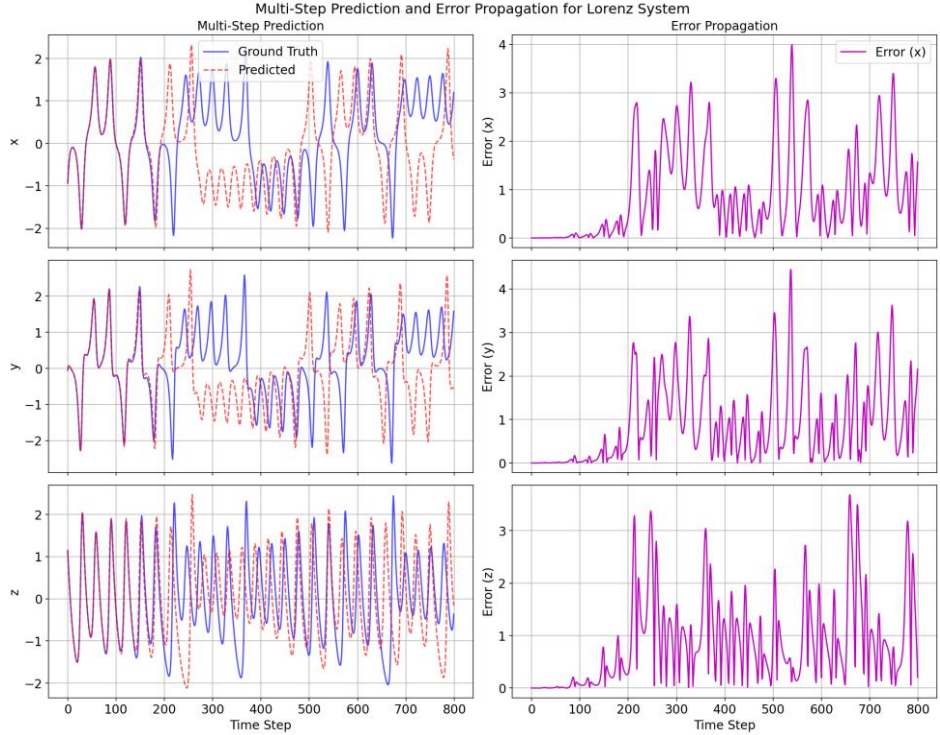

**Fig. 1:** Prediction of Lorenz63 system: The top plot shows, the ground truth and the RFF-RC predicted method during the training and testing phase for one-step ahead prediction. The middle plot shows the butterfly diagram during training and testing phase. The bottom plot shows the multi-step ahead prediction of the Lorenz system.

### 3.1.1 Effect of hyper-parameters

Further, in this study, we analyze the effect of three hyper-parameters: $m$ (the number of random Fourier features), $k$ (the embedding dimension), and $\sigma_{\text{rff}}$ (the Gaussian kernel width) - on model accuracy using normalized root mean square error (NRMSE) as the performance metric. The first plot shows that as $m$ increases, the accuracy improves and then stabilizes, indicating that beyond a certain threshold, additional features do not significantly enhance performance due to saturation in the Monte Carlo approximation process. The second plot highlights the effect of $k$, where the optimal embedding dimension is found to be $k = 3$. This is consistent with the true dimensionality of the Lorenz system. The third plot examines the impact of $\sigma_{\text{rff}}$, with the optimal value determined to be $\sigma_{\text{rff}} = 2$.

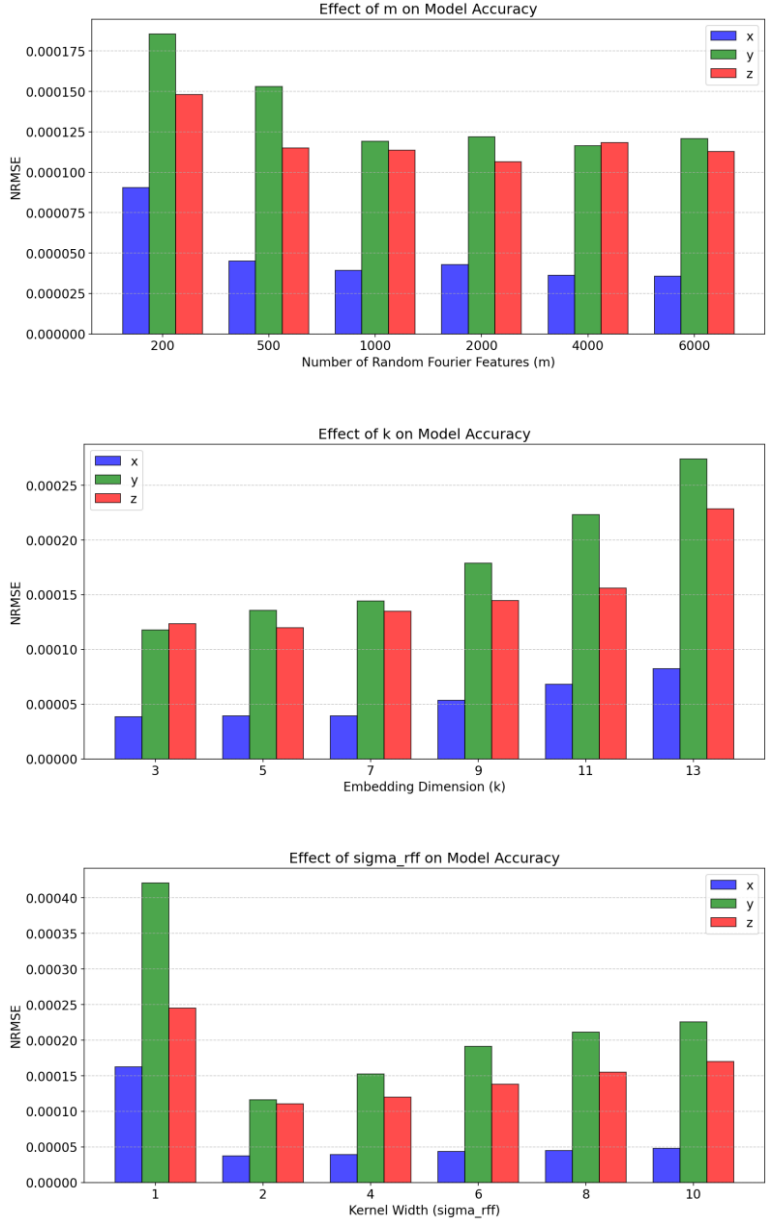

**Fig. 2:** Effect of three hyper-parameters on model accuracy for Lorenz63 system: $m$, the number of random Fourier features (top plot), $k$, the embedding dimension (middle plot), and $\sigma_{\text{rff}}$, the Gaussian kernel width (bottom plot)

### 3.1.2 Noise Robustness Analysis

Figure (3) illustrates the RFF-RC model's performance on the Lorenz63 system under noisy condition. In this analysis we have introduced additive white Gaussian noise (AWGN) at 20 dB to the Lorenz time-series data. The first three subplots compare the noisy ground truth (blue) with the predicted signal (red), highlighting that the model effectively recovers the underlying dynamics despite the noise. The fourth subplot, a 3D phase-space trajectory, further underscores the ability of RFF-RC to reconstruct the characteristic Lorenz attractor even under noisy condition. The test NRMSE values for the $x$, $y$, and $z$ components are $7.17 \times 10^{-3}$, $9.05 \times 10^{-3}$, and $8.70 \times 10^{-3}$, respectively. However, these values are lower than the original one-step-ahead test prediction where only clean data was used to train the model (see Fig. (1)). Still, the method achieves an approximate 15 dB improvement in signal-to-noise ratio (SNR), demonstrating its suitability for denoising chaotic signals while maintaining high predictive accuracy.

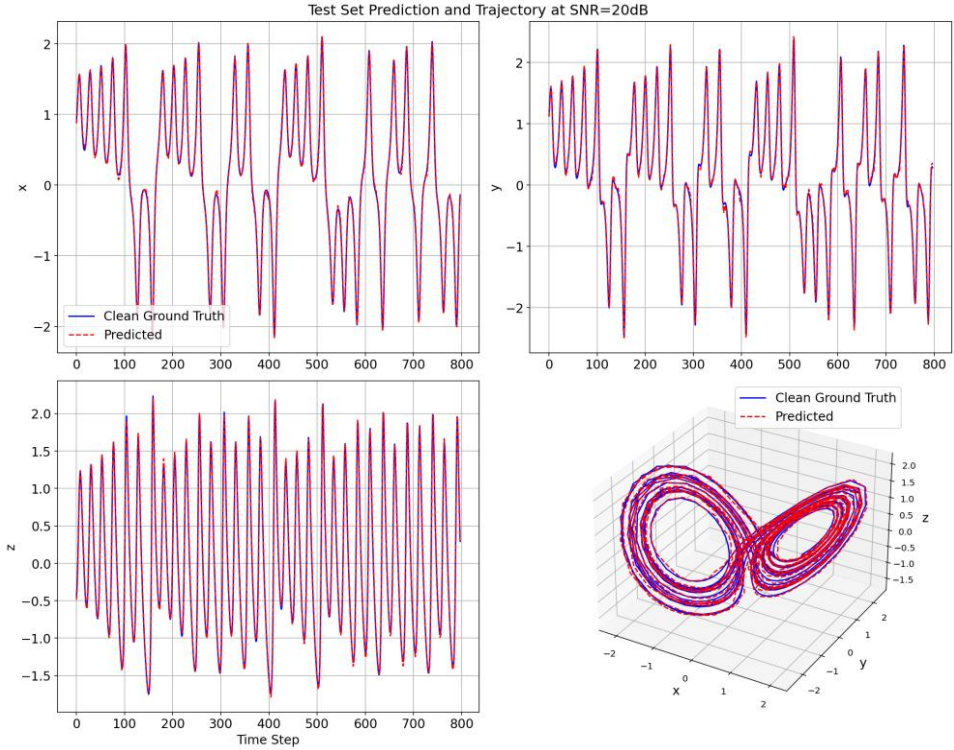

**Fig. 3:** Prediction of Lorenz63 system under noisy condition (20dB SNR) using RFF-RC framework

### 3.1.3 Inferring Lorenz63 System Dynamics from Partial Observations using RFF-RC

In this experiment, we investigate the ability of a Random Fourier Features-based Reservoir Computing (RFF-RC) model to infer the complete dynamics of the Lorenz system using only partial state observations. To achieve this, we first reconstruct the state space from the $x$-component only through delay embedding. This delay-embedded data is then transformed using Random Fourier Features (RFF). Ridge regression is subsequently employed to learn the model weights. This allows the model to predict not only the $x$-component but also the unobserved $y$ and $z$ components. The effectiveness of this approach is evident from the test Normalized Root Mean Squared Error (NRMSE) values, which are $6.85 \times 10^{-5}$ for $x$, $2.24 \times 10^{-4}$ for $y$, and $3.65 \times 10^{-3}$ for $z$. Although the model exhibits the best accuracy for $x$ - the variable it was trained on - the predictions for $y$ and $z$ are surprisingly still robust. This demonstrates that the proposed RFF-RC model can successfully capture the inherent dynamics of the Lorenz attractor even with incomplete input data. When compared to a model trained on the full $x$, $y$, $z$ dataset, where the NRMSE values were $4.08 \times 10^{-5}$, $1.27 \times 10^{-4}$, and $1.19 \times 10^{-4}$ respectively, the increase in error for $y$ and $z$ is expected yet remains within acceptable bounds which underscores the model's efficiency in reconstructing chaotic dynamics from partial measurements. The reconstructed phase-space attractor and time series plots as shown in Fig. (4) further confirm that the predicted trajectories closely follow the ground truth. This indicates that even when trained with limited data, the RFF-RC framework is a powerful tool for inferring full system behavior. This has promising implications for applications in chaotic signal reconstruction and scenarios with limited data.

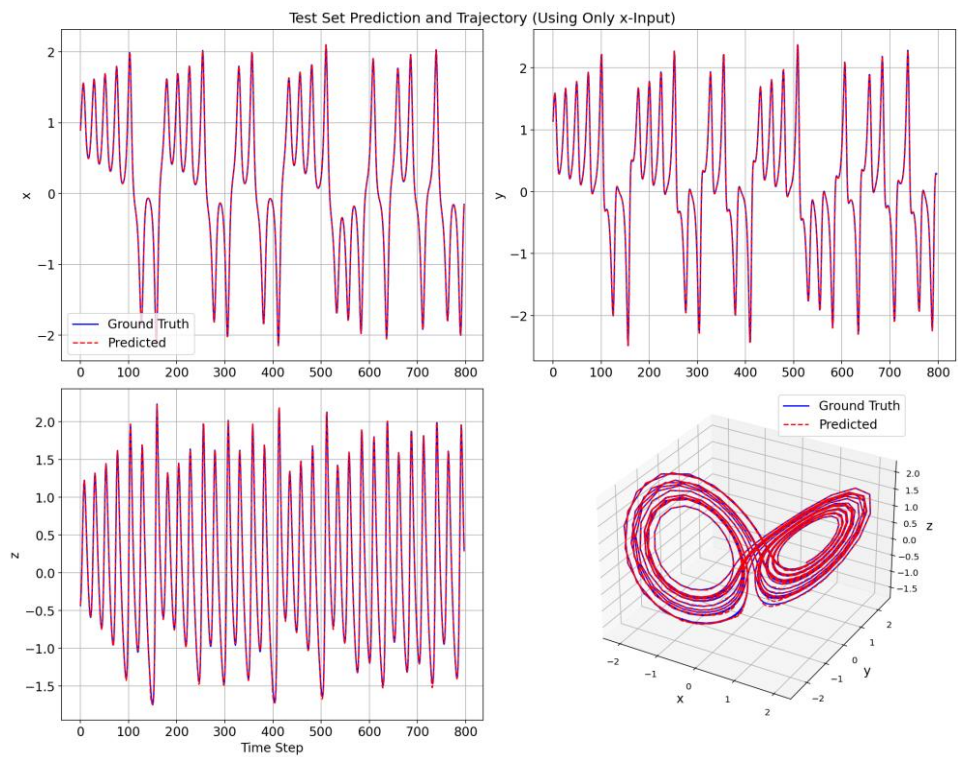

**Fig. 4:** Lorenz attractor reconstruction using partial information (only *x*-data) using RFF-RC framework

## 3.2 Prediction of Mackey-Glass Chaotic Time Series Using Random Fourier Features-Based Reservoir Computing (RFF-RC)

The Mackey-Glass (MG) equation [45] is a well-known time-delay differential equation that exhibits complex chaotic behavior depending on the delay parameter $\tau$. It has been widely used as a benchmark for testing time-series prediction methods due to its intricate temporal dependencies and sensitivity to initial conditions. The delay differential equation is given by

$$\frac{dx}{dt} = \frac{0.2x(t-\tau)}{1+x(t-\tau)^{10}} - 0.1x(t) \qquad (12)$$

In this study, we employ the Random Fourier Features-based Reservoir Computing (RFF-RC) approach to predict the MG time series. We generate a univariate MG time series of 4000 data points with a unit step-size where $\tau = 17$. We again carry out hyper-parameter optimization to determine the best model parameters. The optimal configuration includes a delay embedding dimension $k = 20$, a regularization parameter $\lambda_{\text{reg}} = 1 \times 10^{-8}$, no of Fourier features $m = 4000$, and a kernel bandwidth parameter $\sigma_{\text{rff}} = 2.0$. The performance of the model is evaluated based on Normalized Root Mean Squared Error (NRMSE). The best validation NRMSE is $1.52 \times 10^{-6}$, while training NRMSE is $1.08 \times 10^{-6}$, and one-step-ahead testing NRMSE is $1.97 \times 10^{-6}$. The first plot depicted in Fig. (5) presents the one-step-ahead prediction results, where the predicted time series (magenta) is visually indistinguishable from the actual ground truth (blue), indicating highly accurate short-term forecasting. For multi-step-ahead prediction, we evaluate the performance over 796 steps, achieving an NRMSE of $2.64 \times 10^{-3}$. As shown in the Fig. 5, the model maintains accurate predictions for approximately 500 time steps before deviations appear, which is expected given the chaotic nature of the MG system. The optimal embedding dimension $k = 20$ closely matches the system's delay parameter $\tau = 17$. This aligns with theoretical expectations. These results shown in Fig. (5) demonstrate the effectiveness of RFF-RC in learning the dynamics of a time-delayed system like Mackey Glass equation.

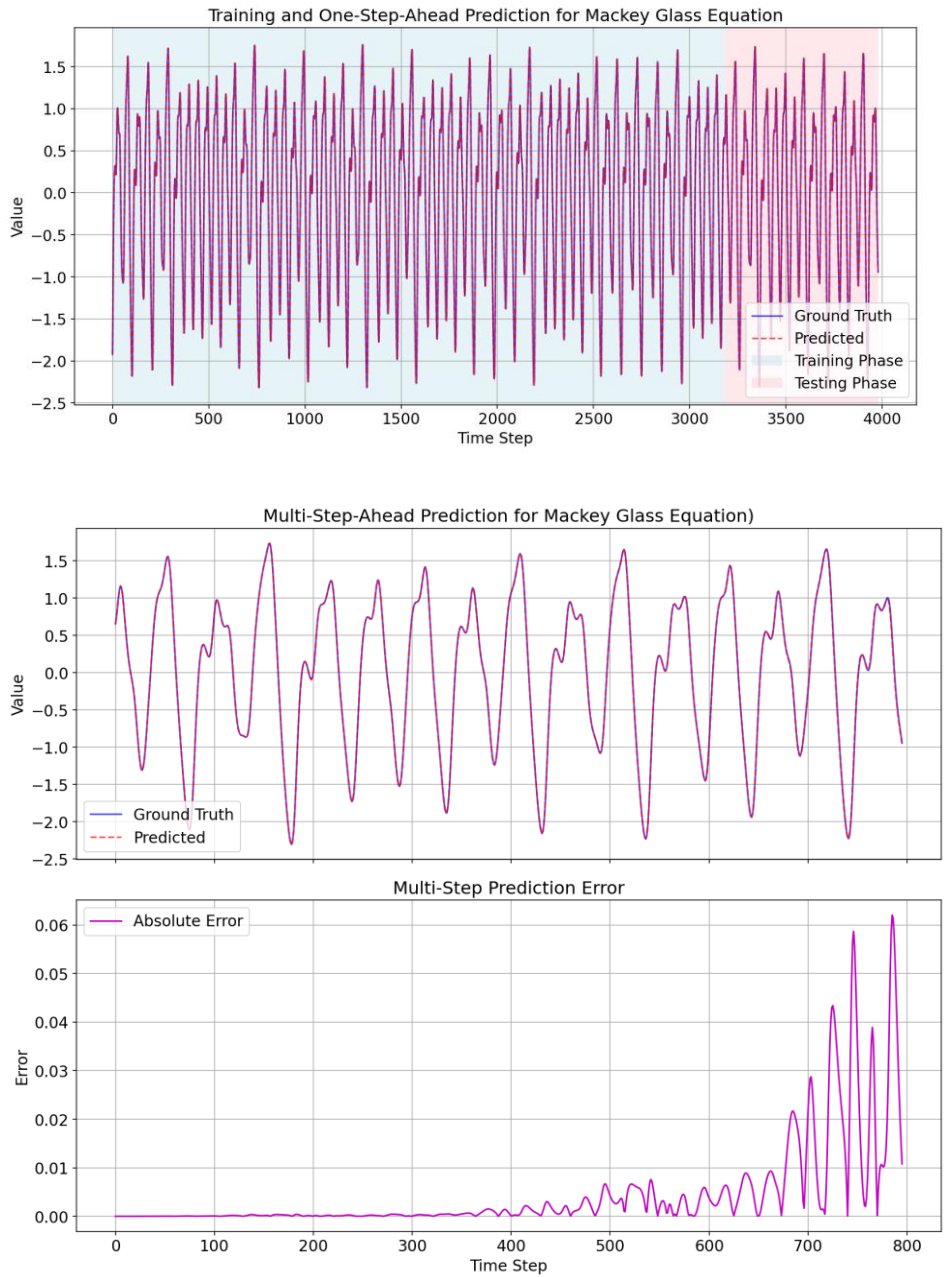

**Fig. 5:** Prediction of Mackey Glass equation: The top plot shows, the ground truth and the RFF-RC predicted method during the training and testing phase for one-step ahead prediction. The bottom plot shows the multi-step ahead prediction of the MG system and the corresponding absolute error.

## 3.3 Prediction of Kuramoto-Sivashinsky (KS) equation Using Random Fourier Features-Based Reservoir Computing (RFF-RC)

In this study, we apply the RFF-RC methodology to the Kuramoto-Sivashinsky (KS) equation [46-47], a canonical nonlinear partial differential equation that models spatiotemporal chaos and pattern formation in spatial dimension. The KS equation is given by

$$u_t + \nu u_{xxxx} + \mu u_{xx} + u u_x = 0 \qquad (13)$$

where $u(x,t)$ represents a scalar field (such as fluid velocity fluctuation or flame front displacement), and $\nu > 0$ and $\mu > 0$ are damping coefficients (we use $\nu = 1$ and $\mu = 1$). The equation also exhibits coherent structures such as traveling waves, unstable periodic orbits, cellular patterns, etc., which serve as topological markers of chaos. Its Lyapunov spectrum features many positive exponents, indicating high-dimensional chaos. And as the system size increases, the KS equation undergoes bifurcations transitioning from low-dimensional chaos to fully developed turbulence, with intermittency characterized by bursts of ordered, quasi-periodic oscillations amid chaotic phases. In our experiments, we consider a spatial domain of $x \in [0, 32\pi]$ with $L = 128$ and apply the RFF-RC framework, which leverages delay embedding and Random Fourier Features to transform complex high-dimensional data into a tractable space for regression. The left-hand side of the Fig. 6 displays the ground truth, one-step ahead prediction, and multi-step ahead prediction (top to bottom), while the right-hand side shows a heatmap of the absolute error during the training phase, testing phase, and multi-step ahead prediction. The results demonstrate that RFF-RC effectively learns and predicts the intricate dynamics of this high-dimensional system. The best parameters obtained after optimization are: $k = 2$, $\lambda_{\text{reg}} = 1 \times 10^{-8}$, $m = 12000$, $\sigma_{\text{rff}} = 20$. It may be noted that, unlike the Mackey-Glass and Lorenz equations, the KS equation necessitates a significantly higher number of Random Fourier Features and a wider kernel bandwidth ($\sigma_{\text{rff}}$). This may be due to the high variance in $u(x,t)$ inherent in high-dimensional spatiotemporal systems.

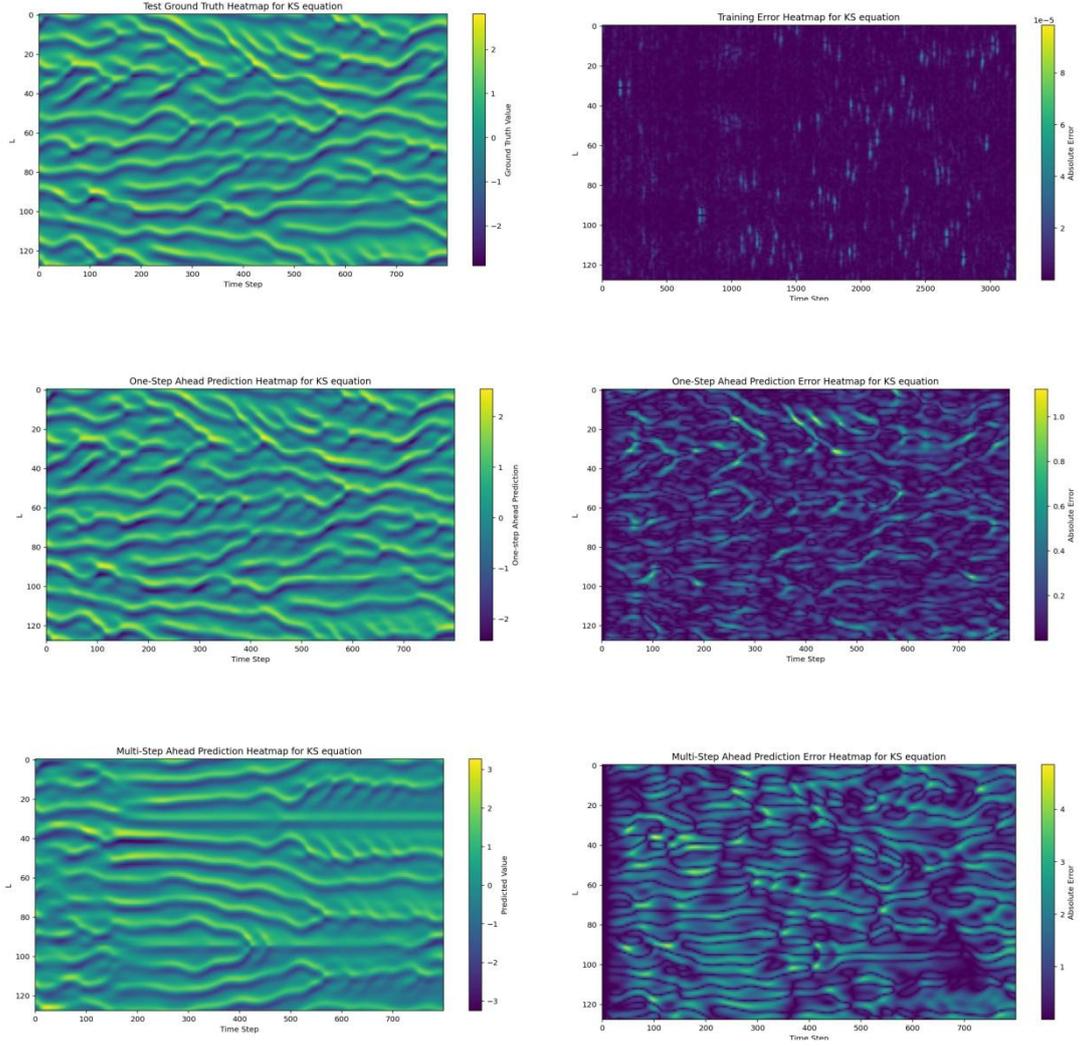

**Fig. 6:** Prediction of Kuramoto-Sivashinsky (KS) equation: The left-hand side of the plot shows the ground truth, one-step ahead prediction, and multi-step ahead prediction (top to bottom), while the right-hand side shows a heatmap of the absolute error during the training phase, testing phase, and multi-step ahead prediction.

## 4. DISCUSSION

In the present work, the integration of delay embedding with Random Fourier Features (RFF) creates a novel reservoir computing framework for forecasting dynamical systems. Delay embedding reconstructs the hidden state from partial observations by stacking $k$ lagged measurements into a delay vector $\mathbf{u}^{\text{delay}}(t) \in \mathbb{R}^{dk}$. Takens' theorem supports this method by ensuring that, for $k > 2d_A$ (with $d_A$ as the attractor's box-counting dimension), the delay-embedded manifold is diffeomorphic to the true state space[42-43]. This preserves key invariants such as Lyapunov exponents and entropy. This also offers deterministic, tunable memory window, $k$. Concurrently, RFF nonlinearly maps these delay vectors into a high-dimensional space $\phi(\mathbf{u}^{\text{delay}}(t)) \in \mathbb{R}^m$ by approximating a kernel, like the Gaussian, through random Fourier sampling. This creates a geometry where linear operations in $\mathbb{R}^m$ mimic nonlinear dynamics, allowing Ridge regression to model $\mathbf{u}(t+1)$ as $\phi(\mathbf{u}^{\text{delay}}(t))\mathbf{W}_{\text{ridge}} + \mathbf{b}_{\text{ridge}}$. This framework couples memory window ($k$) with nonlinear capacity ($m$) and avoids heavy computation. It is robust both theoretically and empirically because Takens' theorem guarantees faithful state reconstruction and kernel trick can handle nonlinearity. The approach works well for partial observations, such as weather data or limited sensor data, by recovering full state dynamics and enriching them with nonlinear features. Ultimately, this RFF-enhanced delay embedding framework provides a scalable, theoretically sound alternative to traditional reservoir computing for chaotic and high-dimensional systems. In essence, RFF creates a feature map by the function $\phi: \mathbf{U} \to \mathbf{F}$ that transforms input data ($\mathbf{u}$) from the input space $\mathbf{U}$ to a feature space $\mathbf{F}$. This transformation is advantageous for many algorithms that require data to be represented in a form suitable for regression or classification.

The key hyper-parameters in the proposed RFF-RC (Random Fourier Features - Reservoir Computing) method include the delay embedding dimension ($k$), the number of Random Fourier Features ($m$), the regularization parameter ($\lambda_{\text{reg}}$), and the kernel width ($\sigma_{\text{rff}}$). Delay embedding dimension ($k$), determines how many past values are used to construct the feature space and capture temporal dependencies. Higher $k$ leads to "over-embedding" and thus lesser accuracy. The regularization parameter, $\lambda_{\text{reg}}$ prevents overfitting by penalizing large model coefficients. The number of Random Fourier Features ($m$) is another crucial hyper-parameter, as it defines the number of basis functions used for feature transformation. Higher $m$ improves approximation at the cost of increased computational complexity. However, after a certain number of $m$, model-predictive accuracy stabilizes. Additionally, the Gaussian kernel bandwidth ($\sigma_{\text{rff}}$) plays a significant role in feature mapping, where a wider bandwidth helps capture broader dependencies but may lead to the loss of finer details. These hyper-parameters collectively influence the model's ability to learn and predict the complex, high-dimensional chaotic dynamics like Mackey-Glass equation, Lorenz equation and Kuramoto-Sivashinsky equation.

## 5. CONCLUSION

We have developed a novel reservoir computing (RC) paradigm that integrates delay embedding with Random Fourier Features (RFF) for forecasting chaotic dynamical systems. Delay embedding reconstructs hidden system dynamics using past observations, ensuring that the delay-embedded manifold remains diffeomorphic to the true attractor, preserving essential system properties. RFF projects these delay vectors into a high-dimensional feature space using randomized Fourier bases, approximating a Gaussian kernel and linearizing complex dynamics without explicit kernel computations. The method relies on only four hyper-parameters: delay embedding dimension, number of RFF, regularization parameter, and kernel bandwidth. Thus, RFF-RC is simpler and more efficient than traditional reservoir computing approaches. We have validated RFF-RC on three canonical chaotic systems: Mackey-Glass, Lorenz, and Kuramoto-Sivashinsky equations. The method achieves high accuracy in one-step-ahead and multi-step-ahead predictions while preserving attractor geometry over long time horizons. The method inherits the noise robustness of delay embedding, smoothing out observational noise through its sliding-window averaging process. Furthermore, it can reconstruct full system dynamics from a single observed variable, leveraging Takens' theorem to bypass the need for full-state measurements. Unlike traditional RC models, RFF-RC is theoretically grounded, stable, and scalable, avoiding gradient-related issues and benefiting from efficient linear algebra operations. This approach bridges dynamical systems theory and machine learning, providing an interpretable, parameter-efficient, and computationally scalable alternative for real-world applications, such as climate modeling and fluid dynamics, where partial data and noisy measurements are common.

**ACKNOWLEDGMENTS**

No funding is received for carrying out this research.

**Conflict of Interest**

The authors have no conflicts to disclose.

**Author Contributions**

**S. K. Laha:** Conceptualization; Formal analysis; Investigation; Methodology; Software; Writing – original draft and reviewing.

## DATA AVAILABILITY

The data that support the findings of this study are available from the corresponding author upon reasonable request.

## REFERENCE

1. Jaeger, Herbert. "The "echo state" approach to analysing and training recurrent neural networks-with an erratum note." *Bonn, Germany: German national research center for information technology gmd technical report* 148.34: 13(2001).
2. Lukoševičius, Mantas, and Herbert Jaeger. "Reservoir computing approaches to recurrent neural network training." *Computer science review* 3, no. 3: 127-149(2009).
3. Lukoševičius, Mantas. "A practical guide to applying echo state networks." In *Neural Networks: Tricks of the Trade: Second Edition*, pp. 659-686. Berlin, Heidelberg: Springer Berlin Heidelberg, 2012.
4. Gauthier, Daniel J. "Reservoir computing: Harnessing a universal dynamical system." *SIAM News51* 12 (2018).
5. Bollt, Erik. "On explaining the surprising success of reservoir computing forecaster of chaos? The universal machine learning dynamical system with contrast to VAR and DMD." *Chaos: An Interdisciplinary Journal of Nonlinear Science* 31, no. 1 (2021).
6. Brunton, Steven L., and J. Nathan Kutz. *Data-driven science and engineering: Machine learning, dynamical systems, and control*. Cambridge University Press, 2022.
7. Brunton, Steven L., Joshua L. Proctor, and J. Nathan Kutz. "Discovering governing equations from data by sparse identification of nonlinear dynamical systems." *Proceedings of the national academy of sciences* 113, no. 15: 3932-3937(2016).
8. Schmid, Peter J. "Dynamic mode decomposition and its variants." *Annual Review of Fluid Mechanics* 54, no. 1: 225-254(2022).
9. Kutz, J. Nathan, Steven L. Brunton, Bingni W. Brunton, and Joshua L. Proctor. *Dynamic mode decomposition: data-driven modeling of complex systems*. Society for Industrial and Applied Mathematics, 2016.
10. Vlachas, Pantelis R., Wonmin Byeon, Zhong Y. Wan, Themistoklis P. Sapsis, and Petros Koumoutsakos. "Data-driven forecasting of high-dimensional chaotic systems with long short-term memory networks." *Proceedings of the Royal Society A: Mathematical, Physical and Engineering Sciences* 474, no. 2213: 20170844(2018).

11. Gilpin, William. "Deep reconstruction of strange attractors from time series." *Advances in neural information processing systems* 33: 204-216(2020).

12. Raissi, Maziar, Paris Perdikaris, and George E. Karniadakis. "Physics-informed neural networks: A deep learning framework for solving forward and inverse problems involving nonlinear partial differential equations." *Journal of Computational physics* 378: 686-707(2019).

13. Li, Zongyi, Nikola Kovachki, Kamyar Azizzadenesheli, Burigede Liu, Kaushik Bhattacharya, Andrew Stuart, and Anima Anandkumar. "Fourier neural operator for parametric partial differential equations." *arXiv preprint arXiv:2010.08895* (2020).

14. Pathak, Jaideep, Brian Hunt, Michelle Girvan, Zhixin Lu, and Edward Ott. "Model-free prediction of large spatiotemporally chaotic systems from data: A reservoir computing approach." *Physical review letters* 120, no. 2: 024102(2018).

15. Pathak, Jaideep, Alexander Wikner, Rebeckah Fussell, Sarthak Chandra, Brian R. Hunt, Michelle Girvan, and Edward Ott. "Hybrid forecasting of chaotic processes: Using machine learning in conjunction with a knowledge-based model." *Chaos: An interdisciplinary journal of nonlinear science* 28, no. 4 (2018).

16. Lu, Zhixin, Brian R. Hunt, and Edward Ott. "Attractor reconstruction by machine learning." *Chaos: An Interdisciplinary Journal of Nonlinear Science* 28, no. 6 (2018).

17. Chattopadhyay, Ashesh, Pedram Hassanzadeh, and Devika Subramanian. "Data-driven prediction of a multi-scale Lorenz 96 chaotic system using deep learning methods: Reservoir computing, ANN, and RNN-LSTM." *Nonlinear Processes in Geophysics Discussions* 2020: 1-26(2020).

18. Pandey, Sandeep, and Jörg Schumacher. "Reservoir computing model of two-dimensional turbulent convection." *Physical Review Fluids* 5, no. 11: 113506(2020).

19. Nakai, Kengo, and Yoshitaka Saiki. "Machine-learning inference of fluid variables from data using reservoir computing." *Physical Review E* 98, no. 2: 023111(2018).

20. Kobayashi, Miki U., Kengo Nakai, Yoshitaka Saiki, and Natsuki Tsutsumi. "Dynamical system analysis of a data-driven model constructed by reservoir computing." *Physical Review E* 104, no. 4: 044215(2021).

21. Chen, Yeyuge, Yu Qian, and Xiaohua Cui. "Time series reconstructing using calibrated reservoir computing." *Scientific Reports* 12, no. 1: 16318(2022).

22. Zimmermann, Roland S., and Ulrich Parlitz. "Observing spatio-temporal dynamics of excitable media using reservoir computing." *Chaos: An Interdisciplinary Journal of Nonlinear Science* 28, no. 4 (2018).

23. Gallicchio, Claudio, Alessio Micheli, and Luca Pedrelli. "Deep reservoir computing: A critical experimental analysis." *Neurocomputing* 268: 87-99(2017).

24. Gauthier, Daniel J., Erik Bollt, Aaron Griffith, and Wendson AS Barbosa. "Next generation reservoir computing." *Nature communications* 12, no. 1: 1-8(2021).

25. Liu, Aodi, Jing Li, Jianfei Bi, Zhangxing Chen, Yan Wang, Chunhao Lu, Yan Jin, and Botao Lin. "A novel reservoir simulation model based on physics informed neural networks." *Physics of Fluids* 36, no. 11 (2024).

26. Hart, Allen, James Hook, and Jonathan Dawes. "Embedding and approximation theorems for echo state networks." *Neural Networks* 128: 234-247(2020).

27. Steil, Jochen J. "Online reservoir adaptation by intrinsic plasticity for backpropagation–decorrelation and echo state learning." *Neural networks* 20, no. 3: 353-364(2007).

28. Schrauwen, Benjamin, Marion Wardermann, David Verstraeten, Jochen J. Steil, and Dirk Stroobandt. "Improving reservoirs using intrinsic plasticity."*Neurocomputing* 71, no. 7-9: 1159-1171(2008).

29. Morales, Guillermo B., Claudio R. Mirasso, and Miguel C. Soriano. "Unveiling the role of plasticity rules in reservoir computing." *Neurocomputing* 461: 705-715(2021).

30. Kawai, Yuji, Jihoon Park, and Minoru Asada. "A small-world topology enhances the echo state property and signal propagation in reservoir computing." *Neural Networks* 112: 15-23(2019).

31. Chouikhi, Naima, Boudour Ammar, Nizar Rokbani, and Adel M. Alimi. "PSO-based analysis of Echo State Network parameters for time series forecasting." *Applied Soft Computing* 55: 211-225(2017).

32. Zhong, Shisheng, Xiaolong Xie, Lin Lin, and Fang Wang. "Genetic algorithm optimized double-reservoir echo state network for multi-regime time series prediction." *Neurocomputing* 238: 191-204(2017).

33. Zhang, Shaohui, Zhenzhong Sun, Man Wang, Jianyu Long, Yun Bai, and Chuan Li. "Deep fuzzy echo state networks for machinery fault diagnosis." *IEEE Transactions on Fuzzy Systems* 28, no. 7: 1205-1218(2019).

34. Rahimi, Ali, and Benjamin Recht. "Random features for large-scale kernel machines." *Advances in neural information processing systems* 20 (2007).

35. Li, Zhu, Jean-Francois Ton, Dino Oglic, and Dino Sejdinovic. "Towards a unified analysis of random Fourier features." In *International conference on machine learning*, pp. 3905-3914. PMLR, 2019.

36. Avron, Haim, Michael Kapralov, Cameron Musco, Christopher Musco, Ameya Velingker, and Amir Zandieh. "Random Fourier features for kernel ridge regression: Approximation bounds and statistical guarantees." In *International conference on machine learning*, pp. 253-262. PMLR, 2017.

37. Sutherland, Danica J., and Jeff Schneider. "On the error of random Fourier features." *arXiv preprint arXiv:1506.02785* (2015).

38. Tanaka, Gouhei, Toshiyuki Yamane, Jean Benoit Héroux, Ryosho Nakane, Naoki Kanazawa, Seiji Takeda, Hidetoshi Numata, Daiju Nakano, and Akira Hirose. "Recent advances in physical reservoir computing: A review." *Neural Networks* 115: 100-123(2019).

39. Yan, Min, Can Huang, Peter Bienstman, Peter Tino, Wei Lin, and Jie Sun. "Emerging opportunities and challenges for the future of reservoir computing." *Nature Communications* 15, no. 1: 2056(2024).

40. Bai, Kang Jun, Clare Thiem, Jack Lombardi, Yibin Liang, and Yang Yi. "Design strategies and applications of reservoir computing: Recent trends and prospects [feature]." *IEEE Circuits and Systems Magazine* 23, no. 4: 10-33(2024).

41. Zhang, Heng, and Danilo Vasconcellos Vargas. "A survey on reservoir computing and its interdisciplinary applications beyond traditional machine learning." *IEEE Access* 11: 81033-81070(2023).

42. Takens, Floris. "Detecting strange attractors in turbulence." In *Dynamical Systems and Turbulence, Warwick 1980: proceedings of a symposium held at the University of Warwick 1979/80*, pp. 366-381. Berlin, Heidelberg: Springer Berlin Heidelberg, 2006.

43. Kantz, Holger, and Thomas Schreiber. *Nonlinear time series analysis*. Cambridge university press, 2003.

44. Lorenz, Edward N. "Deterministic Nonperiodic Flow 1." In *Universality in Chaos, 2nd edition*, pp. 367-378. Routledge, 2017.

45. Mackey, Michael C., and Leon Glass. "Oscillation and chaos in physiological control systems." *Science* 197, no. 4300: 287-289 (1977).

46. Kuramoto, Yoshiki. "Diffusion-induced chaos in reaction systems." *Progress of Theoretical Physics Supplement* 64: 346-367 (1978).

47. Ashinsky, Gi Siv. "Nonlinear analysis of hydrodynamic instability in laminar flames—I. Derivation of basic equations." In *Dynamics of Curved Fronts*, pp. 459-488. Academic Press, 1988.